\pdfoutput=1

\documentclass[11pt, dvipsnames]{article}

\usepackage[table]{xcolor}
\usepackage[final]{acl}

\usepackage{graphicx}
\usepackage{amsmath}

\usepackage{booktabs}
\usepackage{array,multirow,graphicx}
\usepackage{float}
\usepackage{makecell}
\usepackage{amssymb}
\usepackage{pifont}

\usepackage{algorithm}
\usepackage{algorithmic}
\usepackage{hyperref}
\usepackage{multirow}
\usepackage{multicol}
\usepackage{graphics}
\usepackage{array}
\usepackage{array, diagbox}

\usepackage{booktabs}
\usepackage{arydshln}
\usepackage{multirow}
\usepackage{times}
\usepackage{latexsym}
\usepackage[multiple]{footmisc}
\definecolor{rowgray}{gray}{0.95}

\usepackage{times}
\usepackage{latexsym}

\usepackage[T1]{fontenc}

\usepackage[utf8]{inputenc}

\usepackage{microtype}

\usepackage{inconsolata}

\usepackage{xcolor}
\usepackage{titling}

\title{Predicting Causal Effects from Natural Language Queries using Structured Representations}

\author{%
    Giuliano Martinelli$^1$, Piriyakorn Piriyatamwong$^1$, Abelardo Carlos Martinez Lorenzo$^1$,\\
    \textbf{Jasmin Baier$^{1,2}$}, \textbf{Riccardo Orlando$^1$}, \textbf{Satvik Garg$^1$}, \textbf{Sharif Kazemi$^1$},  \textbf{Linxi Wang$^1$},\\ 
    \textbf{Arianna Legovini$^1$}, \textbf{Samuel P. Fraiberger$^{1,3}$} \\
    $^1$The World Bank Group, $^2$University of Oxford, $^3$New York University\\
\texttt{\{gmartinelli, ppiriyatamwong, amartinezlorenzo, jbaier, rorlando,} \\ \texttt{sgarg8, msharifkazemi, lwang34, alegovini, sfraiberger\}@worldbank.org}
}
\date{} 

\begin{document}
\maketitle

\vspace{-0.8em}

\begin{abstract}
Randomized controlled trials are a cornerstone of medicine and the social sciences as they enable reliable estimates of causal effects. 
However, they are costly and time-consuming to conduct, motivating interest in predicting causal effects from existing experimental evidence. 
Recent advances in large language models (LLMs) have demonstrated strong performance on knowledge-intensive tasks, raising the question of whether these models can be used for forecasting causal effect sizes.
To investigate this, we introduce Query2Effect, a new large-scale benchmark consisting of more than 72,000 natural language questions aligned with experiment descriptions, created to simulate realistic information-seeking scenarios by varying query specificity along dimensions of implicitness, abstraction, and ambiguity. We then propose a two-step framework that first generates a synthetic structured representation of a query before predicting effect size using a supervised encoder model. Experiments show that finetuning plays a crucial role in improving prediction performance, with absolute error reducing by -27\% up to -71\% compared to prompted out-of-the-box LLMs, and that our two-step framework is beneficial for out-of-domain generalization, highlighting the benefits of separating semantic interpretation from numerical effect estimation.
\end{abstract}

\section{Introduction}\label{sec:introduction}

\begin{figure}[t!]
    \centering
    \includegraphics[width=0.98\linewidth]{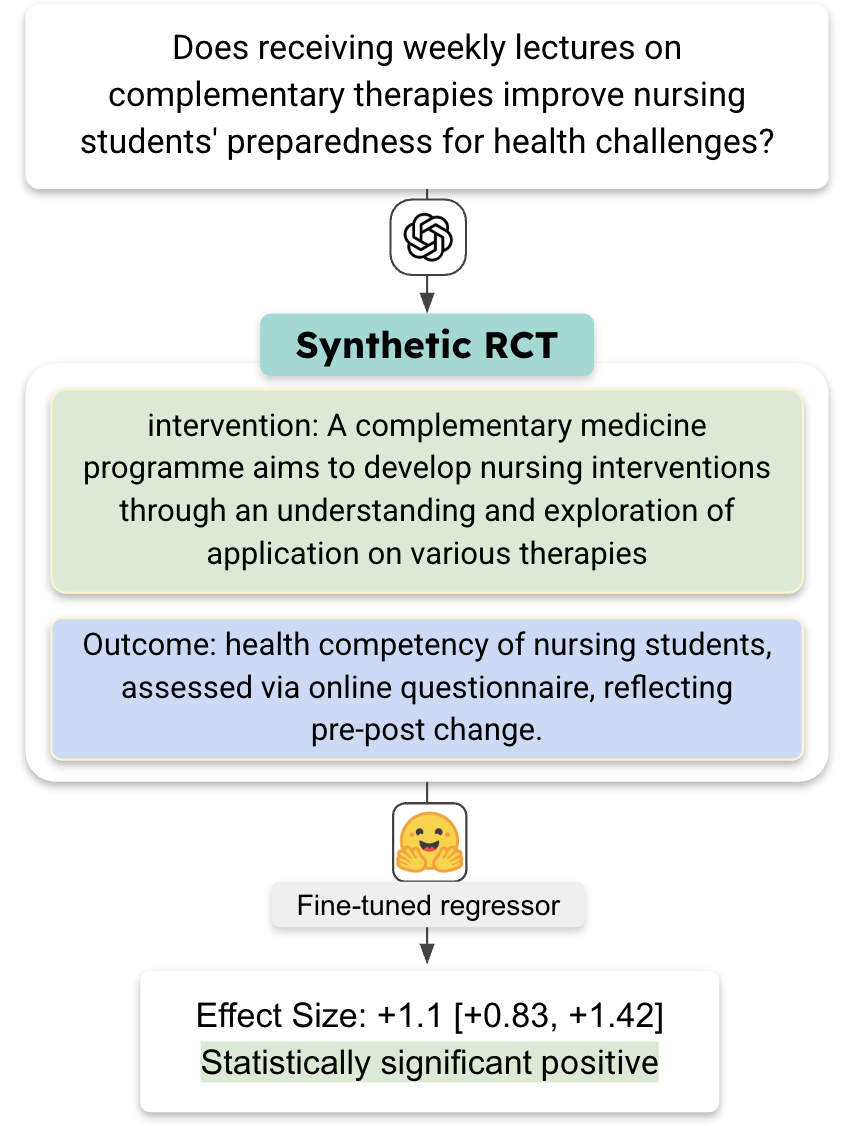}
    \caption{Overview of the Synthetic-RCT pipeline. A natural language query is first mapped to a structured representation of an intervention-outcome relationship, after which a specialized model predicts the corresponding effect size.}
    \label{fig:example}
\end{figure}

Decision makers are often confronted with causal questions, such as: \textit{“By how much can fertilizer vouchers increase maize yields for smallholder farmers in rural Africa?”}. Randomized controlled trials (RCTs) provide answers to such questions by empirically estimating the causal effect of an intervention (e.g. \textit{fertilizer}) on an outcome of interest (e.g. \textit{maize yields}), and have therefore become the cornerstone of evidence-based decision making in the medical and social sciences ~\cite{duflo2017handbook}. However, running such experiments in real-world settings is costly and time-consuming, and many practical questions arise before direct evidence is available.

Recent advances in natural language processing offer new possibilities for addressing this challenge. Large language models (LLMs) have demonstrated strong performance on knowledge-intensive tasks, raising the question of whether they can capture statistical regularities across textual descriptions of existing causal studies that are predictive of effect sizes. Most of the NLP literature related to scientific evidence focuses on retrospective tasks such as retrieval \cite{acharya-etal-2025-m3retrieve}, structured extraction \cite{lasri-etal-2023-econberta}, and evidence synthesis \citep{ramprasad-etal-2023-automatically, wang2025accelerating}. Existing work on forecasting is typically framed as a discrete classification task (e.g., positive vs. negative effect) and assumes access
to detailed experimental descriptions~\citep{luo2025large, kolluri2025finetuning}. These approaches therefore do not address quantitative effect-size prediction from underspecified natural-language queries and are often evaluated on stylized laboratory settings rather than heterogeneous real-world field interventions~\cite{hewitt2024predicting, anthis2025llm}. 

In this paper, we study a more challenging problem: \textbf{Can we predict continuous causal effect sizes directly from textual queries?} This task is \textit{scientifically relevant} because it tests whether models can accurately estimate quantitative treatment effect information from language alone, and \textit{practically relevant} because such forecasts could support evidence-based decision making in domains such as public policy and medicine. 

A central obstacle is the lack of benchmarks linking free-form textual queries to causal effects. 
To address this gap, we introduce \textbf{Query2Effect}, a new large-scale benchmark constructed via a reverse supervision strategy. 
Starting from a curated corpus of randomized controlled trial descriptions and associated effect sizes, we generate more than 72{,}000 natural language queries from their intervention-outcome descriptions. 
The dataset systematically varies query specificity along dimensions of implicitness, abstraction, and ambiguity to reflect the range of questions encountered in realistic scientific and applied settings, enabling controlled evaluation under different levels of linguistic underspecification.
We further propose the \textbf{Synthetic-RCT} pipeline, a modular framework that first reconstructs a structured representation of the intervention-outcome relationship implied by the query, and then predicts the corresponding effect size (Figure~\ref{fig:example}). 
This design explicitly separates semantic interpretation from effect estimation, allowing the model to use a structured experimental abstraction as an intermediate representation.

Our experiments show that out-of-the-box prompted LLMs perform poorly at predicting effect sizes from natural language queries. Supervised models trained on our benchmark substantially improve predictive accuracy, and incorporating a structured intermediate representation yields further gains in robustness and out-of-domain generalization.

To summarize, in this paper, we put forward:
\begin{itemize}
    \item \textbf{A new NLP task} of predicting continuous causal effect sizes from natural language queries.
    \item \textbf{Query2Effect}, a large-scale benchmark linking over 72{,}000 natural language queries to causal effect estimates derived from randomized controlled trials.
    \item \textbf{The Synthetic-RCT pipeline}, a structured modeling approach that explicitly separates semantic interpretation from numerical prediction and substantially improves over prompted out-of-the-box LLM baselines.
\end{itemize}

\section{Related Work} \label{sec:related-works}

\subsection{Predicting Experimental Outcomes with LLMs}

A growing body of work investigates whether LLMs can predict the outcomes of scientific and behavioral experiments. 
In neuroscience, \citet{luo2025large} show that models trained on domain literature can predict experimental results with accuracy exceeding that of human experts on benchmark tasks. 
In the social and behavioral sciences, LLMs have been used to simulate survey responses and approximate aggregate experimental effects.
\citet{hewitt2024predicting} demonstrate that LLM outputs correlate with observed effects in survey experiments, while \citet{kolluri2025finetuning} report improved behavioral realism when LLMs are fine-tuned on large collections of human responses. 
Other recent studies find that LLMs can approximate human decision-making in controlled settings and support exploratory social simulations \citep{guo-etal-2025-estimating, meister-etal-2025-benchmarking, anthis2025llm}. 
Despite these advances, most existing approaches focus on predicting individual or aggregate responses rather than quantitative treatment effects from real-world field experiments.  
A closely related work by \citet{chenpredicting} evaluates LLM predictions across economic field experiments, showing that models struggle to correctly interpret interventions and generalize across contexts. 
However, their evaluation is limited to multiple-choice question answering using out-of-the-box LLMs. 
By contrast, we study the more realistic task of predicting continuous treatment effects from underspecified natural language queries, introducing a large-scale benchmark grounded in experimental evidence.

\subsection{Human Forecasting of Treatment Effects} \label{sec:related-works-human}

A parallel literature examines whether humans can predict the results of experiments. \citet{vivalt2020much}, \citet{dellavigna2022rcts}, and \citet{dellavigna2025forecasting} show that expert predictions contain signal but are significantly biased, typically overestimating experiment outcomes. 
While aggregated forecasts better correlate with realized outcomes, individual-level accuracy remains limited. 
These findings highlight the intrinsic difficulty of extrapolating treatment effects across contexts, even for domain experts, and motivate the search for computational approaches that can leverage information from prior studies. 

\subsection{Structured Representations for Causal Reasoning}

Early work demonstrates the potential of using textual descriptions to predict causal relations \cite {chen2024proximal, bansal-sharma-2023-controlling}.
However, more recent studies in causal representation learning suggest that directly estimating causal effects from unstructured text may produce unstable or biased predictions \cite{corcoll2024contrastive}. To mitigate this, these approaches aim to isolate causally relevant information from text through contrastive objectives or latent variable models that disentangle treatment semantics from confounding variation \citep{imai2024causal_rep_learning, imai2025genai}. 
More broadly, a related line of work shows that structured or abstract intermediate representations improve generalization across domains and tasks. 
In applied settings, structured extraction frameworks such as PICO decompose experimental descriptions into key components such as population, intervention, comparator, and outcome using neural and large language models \citep{wadhwa2023jointly, ghosh2024alpapico}. 
In representation learning, models that learn abstract latent factors from entangled inputs generalize more efficiently to unseen settings \citep{johnston2023abstract, hummos2024gradient}.
From a theoretical perspective, domain generalization frameworks grounded in the information bottleneck formalize how compressed invariant representations improve robustness to distribution shift \citep{zhang2023deep}. 
These findings motivate approaches that separate semantic abstraction from numerical estimation. 
Our Synthetic-RCT framework follows this principle by generating structured intermediate representations of interventions--outcome relationships, serving as a semantic bottleneck before predicting treatment effect sizes.

\section{Query2Effect Dataset}
\label{sec:dataset}

In this work, we construct the first large-scale benchmark linking free-form queries to causal effect estimates. 
We built this dataset from a curated corpus of RCT studies paired with structured effect-size information.\footnote{\scriptsize{\url{https://impactai.worldbank.org/blog/extracting-experimental-data-from-studies}}} 
From these studies, we generate multiple natural language queries at varying levels of ambiguity and abstraction using a new taxonomy of query specificity.

\subsection{Source Data and Preprocessing}
\label{sec:dataset-corpus}

\begin{table*}[t]
\centering
\resizebox{\linewidth}{!}{
\begin{tabular}{p{3cm}  p{8cm}  p{12cm}}
\toprule
\textbf{Data Field} & \textbf{Description} & \textbf{Example} \\
\hline
\rowcolor{gray!20}
\multicolumn{3}{c}{Example drawn from RCT dataset. } \\
\hline
Estimate id & Unique id of this intervention outcome pair & 76717 \\
\hline
Intervention Descr. & Detailed description of intervention & \textit{The intervention involves the introduction of malaria rapid diagnostic tests (mRDTs) in public health centers, where healthcare personnel diagnose malaria using mRDTs instead of relying solely on clinical judgment. ACT is only administered when the mRDT result is positive.}\\
\hline
Outcome Descr. & Detailed description of outcome & \textit{Aggregate societal cost (health sector + household) per 1000 fever episodes over 2 years.}\\
\hline
Effect Size & standardized effect size & [-0.0129] \\
\hline
Confidence Inter. & 95\% confidence intervals & [-0.101, 0.075]\\
\hline
\rowcolor{rowgray}
\multicolumn{3}{c}{Elements created using our taxonomy from original RCTs}\\
\hline
Query Level-0 & Implicitness:I0, Abstraction:A0, Ambiguity:U0 & \textit{What is the effect of introducing malaria rapid diagnostic tests (mRDTs) in public health centers for diagnosing malaria in children under five in rural Ghana, compared to relying solely on clinical judgment, on the aggregate societal cost per 1000 fever episodes over two years?}\\
\hline
Query Level-1 & Implicitness:I1, Abstraction:A1, Ambiguity:U1 & \textit{What impact does the introduction of mRDTs for malaria diagnosis in public health centers have on the societal costs associated with managing fever cases?}\\
\hline
Query Level-2 & Implicitness:I2, Abstraction:A2, Ambiguity:U2 & \textit{How does introducing a new diagnostic tool in healthcare settings affect resource utilization efficiency compared to traditional diagnostic methods?}\\
\hline
Query Level-3 & Implicitness:I3, Abstraction:A3, Ambiguity:U3 & \textit{How do diagnostic advancements influence public health economics?}\\
 \bottomrule
\end{tabular}
}
\caption{Overview of the dataset structure. The top panel shows an example RCT entry with its data fields and descriptions. The bottom panel shows four corresponding queries generated at different levels of specificity.}
\label{tab:queries}
\end{table*}

Our starting point is an expert-curated corpus of 7,354 RCTs spanning multiple sectors, including health, education, agriculture, public finance, and social protection. Because individual trials often report multiple intervention–outcome pairs, the repository provides 74,826 causal effect estimates.

For each estimate $EST_i$, the dataset provides intervention description $I_i$, outcome description $O_i$, standardized causal effect size $ES_i$, and the lower and upper bounds of the 95\% confidence interval $[CL_i, CU_i]$. Table~\ref{tab:queries} presents these components,  along with an illustrative example. 
Additional details on the corpus and effect size standardization are provided in Appendix \ref{app-rct}.

From this corpus, we retain only estimates corresponding to a single intervention and a single outcome, focusing on settings where the effect of an intervention is evaluated against a status quo or no-treatment baseline. 
This mirrors the structure of typical natural language queries that rarely specify multiple competing alternatives.
As recent literature highlights that health RCTs have lower heterogeneity in intervention and outcome definitions, higher reporting standardization, and better domain consensus \cite{Hopewell2025, Upsher2025}, we build our main training-validation-test splits focusing exclusively on the health sector.
To evaluate cross-domain performance, we reserve an additional out-of-domain testset, Test$_{ood}$, composed of non-health RCTs including estimates of the other sectors such as education, agriculture, and social protection, inter alia.
The resulting corpus contains 13,804 intervention–outcome estimates for our main health-related dataset, and 4,227 estimates for the out-of-domain testset, drawn from a total of approximately 2,700 RCTs.
While this final corpus contains precise experimental descriptions, it does not reflect how domain experts typically formulate questions in practice. 
Instead of structured variables, practitioners often express information needs through natural language, such as questions or problem statements. 
In the following section, we bridge this gap by introducing a taxonomy of query specificity, which we use to generate linguistically diverse queries that preserve the underlying causal semantics while varying in implicitness, abstraction, and ambiguity.

\subsection{Query Specificity}
\label{sec:dataset-taxonomy}

In practice, natural-language queries vary in how precisely they specify the underlying causal problem, as users may omit key elements of the study design or abstract away from operational details. 
To capture this variability, we introduce a taxonomy of query specificity along three orthogonal linguistic dimensions:
\begin{enumerate}
    \item \textbf{Implicitness} (I): degree to which causal elements are explicitly stated.
    \item \textbf{Abstraction} (A): level of conceptual generalization in phrasing.
    \item \textbf{Ambiguity} (U): degree of underspecification and interpretive uncertainty.
\end{enumerate}
Each dimension is discretized into four ordered levels, producing a structured space that ranges from very specific concrete queries to highly abstract and ambiguous formulations. Table~\ref{tab:taxonomy} summarizes this taxonomy.
This design serves two purposes. 
First, it enables controlled generation of linguistically diverse queries while preserving the underlying causal semantics. 
Second, it allows us to study how varying levels of semantic compression and underspecification affect effect size prediction. 

\begin{table}[t]
\centering
\small
\begin{tabular}{lll}
\toprule
\textbf{Dimension} & \textbf{Level} & \textbf{Description} \\
\midrule
Implicitness & I0 & All causal elements explicit \\
             & I1 & One element implicit \\
             & I2 & Multiple elements abstracted \\
             & I3 & Most elements implicit \\
\midrule
Abstraction  & A0 & Concrete phrasing \\
             & A1 & Paraphrased concrete \\
             & A2 & Conceptual abstraction \\
             & A3 & High-level framing \\
\midrule
Ambiguity    & U0 & Clear causal intent \\
             & U1 & Mild underspecification \\
             & U2 & Multiple interpretations \\
             & U3 & Ill-posed or vague \\
\bottomrule
\end{tabular}
\caption{Query specificity taxonomy, illustrating the dimensions of Implicitness, Abstraction, and Ambiguity and their corresponding levels.}
\label{tab:taxonomy}
\end{table}

\subsection{Automatic Query Generation}
\label{sec:dataset-generation}
For each RCT estimate $EST_i$, we generate multiple queries that preserve the underlying causal semantics while varying linguistic specificity.

Specifically, we use Gemini-3-pro~\footnote{\scriptsize{\url{https://ai.google.dev/gemini-api/docs/gemini-3}}} to generate queries corresponding to different configurations of the query specificity taxonomy. 
For each estimate, we generate four queries spanning the $(I, A, U)$ dimensions (Table~\ref{tab:taxonomy}), ranging from fully specified formulations to increasingly abstract or underspecified queries, while referring to the same underlying intervention-outcome relationship. 
Each query is constrained to: (i) preserve semantic consistency with $EST_i$, (ii) avoid introducing information not supported by the trial description, (iii) refrain from hallucinating experimental details, and (iv) be expressed as a single sentence. Prompt templates and the full generation protocol with descriptions of the four query specificity levels are provided in Appendix~\ref{app:prompts-query}.

\subsection{Human validation}
We perform human validation to evaluate two complementary dimensions of the generated queries: (i) \textit{faithfulness}, i.e., whether they preserve the causal meaning of the original RCT estimates while respecting the intended specificity constraints, and (ii) \textit{realism}, i.e., whether they are close to the style and distribution of those that would naturally be asked by expert economists or policymakers.
We ask two human annotators to assess a subset of 500 generated queries independently. 
For each query $Q_i$ generated from the corresponding estimate $EST_i$, annotators evaluate whether the query was semantically faithful to the underlying RCT and whether the query appears realistic and professionally formulated, resembling the type of question an expert practitioner might ask when seeking evidence about the expected impact of a treatment or policy intervention.
The results indicate that 96\% of the generated queries were judged to be semantically faithful to the original RCT estimates, and 94\% of the queries were judged to be realistic expert-like questions with near-perfect agreement between the two annotators.

As an additional evaluation of realism, we conduct a blind A/B test. 
First, one annotator manually produced 2,000 queries based solely on RCT intervention and outcome information. 
Then, we construct 2,000 query pairs, each containing one human-written query and one automatically generated query, with well-balanced distribution between specific and unspecific queries.
A second annotator, blind to the source of each query, is then asked to identify which query in each pair had been written by a human expert.
The annotator selected the correct query in only 52\% of cases, a result close to random chance, suggesting that the automatically generated queries are difficult to distinguish from human-written ones and providing further evidence that the generation pipeline produces high-quality, realistic queries.

\subsection{Final Dataset Statistics} 
\label{sec:dataset-stats}
The final Query2Effect dataset contains 72,000 query–effect pairs. 
We split our health-related dataset into standard train, validation, and testing splits, ensuring that estimates originating from the same RCT do not appear in multiple splits, preventing information leakage. 
Summary statistics of our final dataset, including details about our in-domain and out-of-domain testsets, Test$_{id}$ and Test$_{ood}$, are reported in Table~\ref{tab:data_stats}.


\begin{table}[t]
\centering
\small
\begin{tabular}{ccccc}
\toprule
\textbf{ } & \textbf{Train} & \textbf{Val} & \textbf{Test$_{id}$}  & \textbf{Test$_{ood}$} \\
\midrule
RCTs & 10,467 & 1,275 & 2,062  & 4,227\\
Queries & 41,868 & 5,100 & 8,248  & 16,908\\
Query lvl-0  & 222 & 225 & 224  & 238\\
Query lvl-1  & 135 & 139 & 138  & 135\\
Query lvl-2  & 107 & 109 & 108  & 131\\
Query lvl-3  & 105 & 106 & 106  & 98\\
\bottomrule
\end{tabular}
\caption{We report the total number of RCTs and queries for each split, along with the average character length of queries at different levels of specificity.}
\label{tab:data_stats}
\end{table}

\section{Task and Methods}
\label{sec:methods}

\subsection{Problem Formulation}
\label{sec:problem}

We study the problem of \textit{text-based causal effect forecasting}. Given a natural language query $q$ describing a potential intervention and outcome 
, the objective is to predict the associated standardized effect size with confidence intervals:
$$
f_\theta(q_i) = (\hat{E_i}, \hat{CL_i}, \hat{CU_i}),
$$
where $q_i$ is the query and $\hat{E_i}$, $\hat{CL_i}$ and $\hat{CU_i}$ denote predicted standardized effect size estimates and confidence bounds.
This setting differs from classical causal inference, as the model (i) does not observe treatment or control outcomes, (ii) has no access to covariates or full experimental data, and (iii) must generalize across heterogeneous interventions and populations.
The task is therefore one of semantic extrapolation: the model must infer likely causal magnitude from language alone, using distributional patterns learned from pretraining or additional in-domain finetuning. 


\subsection{Modeling Approaches}


\subsubsection{End-to-End Forecasting}

In the end-to-end setting, a model directly predicts the treatment effect and its confidence interval from a natural language query.
This formulation requires the model to jointly interpret the intervention and outcome semantics, implicitly reconstructing the experimental structure underlying the query.

\subsubsection{Synthetic-RCT Pipeline}
We propose to decompose the prediction task into two stages, by first predicting a structured RCT representation, and then estimating the effect size. We define a synthetic RCT representation as two minimal descriptions of the intervention and the outcome.
Although this abstraction is deliberately simple, we hypothesize that these two elements capture the core causal relationship underlying most RCT evidence. 
The synthetic RCT representation acts as a \textit{semantic bottleneck}, mirroring how a human might reason: first identifying the relevant intervention–outcome relationship, and only then recalling or estimating the magnitude of the effect.

Given a query $q$, a language model extracts or reconstructs this structure, while avoiding hallucinating experimental details or introducing unsupported quantitative information. This structured representation is then linearized into text and provided as input to the effect prediction model. 
We detail exact prompts used for this step in Appendix~\ref{app:prompts-es}.

\section{Experimental Setup}
\label{sec:exp_setup}
\subsection{Datasets}
We conduct our experiments on two datasets: the Query2Effect dataset, which we use to train models and evaluate in- and out-of-domain capabilities, and AidGrade$_{ood}$, a dataset adapted from prior work in development economics \cite{vivalt2020much}, to evaluate out-of-domain performances.

\paragraph{Query2Effect.}
Our primary evaluation is based on our new benchmark, Query2Effect. We use the training and validation splits of Query2Effect to train our supervised systems, and evaluate model performance on the two different testsets: Test$_{id}$, to benchmark performance on in-domain health-related queries, and Test$_{ood}$, to benchmark performance on out-of-domain cross-sector queries.


\paragraph{AidGrade$_{ood}$.}
To further assess generalization performance, we construct a secondary benchmark based on the meta analysis dataset AidGrade\footnote{\scriptsize\url{http://www.aidgrade.org/get-data}} \cite{vivalt2020much}. 
This dataset aggregates evidence from more than 600 RCTs across development economics interventions in multiple sectors, such as education and farming. 
As the source reports aggregate estimates for intervention-outcome pairs, we construct a smaller out-of-domain dataset of 73 queries with a templatic form, in which each effect size reflects a more stable aggregate estimate.
We use Appendix \ref{app:ood} to report the detailed process to construct this dataset.




\subsection{Evaluation Metrics}

We evaluate model performance along two dimensions: 
(i) \textit{statistical accuracy} in predicting continuous effect sizes and 
(ii) \textit{policy relevance}, capturing whether models correctly identify the direction and practical importance of treatment effects.

\paragraph{Regression Metrics.}
We measure the numerical accuracy of predicted effect sizes using standard regression metrics: Root Mean Squared Error (RMSE), Mean Absolute Error (MAE), the coefficient of determination ($R^2$), and correlation-based metrics, including Pearson and Spearman correlation. 
These metrics evaluate both the magnitude of errors and the ability to preserve the relative ordering of treatment effects.

\paragraph{Policy-Oriented Metrics.}
Because policy decisions often rely on qualitative judgments about intervention impact, we additionally report decision-oriented metrics. 
Direction Accuracy measures whether predicted and true effects share the same sign. 
Economic Significance evaluates whether models correctly identify practically meaningful effects. 
Finally, Statistical Significance evaluates whether models correctly classify effects as positive, negative, or non-significant based on their 95\% confidence intervals.
Formal definitions are provided in Appendix~\ref{app-metrics}.

\begin{table*}[ht!]
\centering
\resizebox{\linewidth}{!}{
\begin{tabular}{l c c c c c c c c}
\toprule
\textbf{Model} & \textbf{RMSE$\downarrow$} & \textbf{MAE$\downarrow$} & \textbf{$R^2$$\uparrow$} & \textbf{Pearson$\uparrow$} & \textbf{Spearman$\uparrow$} & \textbf{Dir. $\uparrow$} & \textbf{Econ. Sign. $\uparrow$} & \textbf{Stat. Sign. $\uparrow$}
\\

\toprule
\textit{mean-effect} & \textit{0.2788} & \textit{0.1946} & \textit{-0.0797} & - & - & \textit{73.02} & \textit{48.46} & \textit{45.3} \\
\textit{retrieval-lookup} & \textit{0.3880} & \textit{0.2526} & \textit{-0.9235} & \textit{0.2423} & \textit{0.164} & \textit{67.8} & \textit{51.0} & \textit{53.3}
\\

\hline
\rowcolor{gray!20}
\multicolumn{9}{c}{\textit{Prompted}} \\
Gemini 2.5 Flash & 0.4503 & 0.3542 & -1.5908 & 0.1711 & 0.2014 & 59.36 & 51.36 & 41.37
\\
GPT-OSS 120B & 0.4900 & 0.4160 & -2.0682 & 0.1977 & 0.2213 & 59.80 & 48.88 & 32.98
\\
GPT-5.2 & 0.3165 & 0.2374 & -0.2801 & 0.2700 & 0.3008 & 65.23 & 52.33 & 62.22 
\\


\hline
\rowcolor{gray!20}\multicolumn{9}{c}{\textit{Our models (Supervised)}} \\

ModernBERT$_q$ & 0.2659 & 0.1743 & 0.097 & 0.348 & 0.2988 & 71.82 & 57.71 & 66.78 
\\
Synthetic-RCT(\small GPT-OSS-20B\large) $\rightarrow$ ModernBERT$_r$ & 0.2685 & 0.179 & 0.0807 & 0.3349 & 0.3002 & 72.63 & 57.71 & 65.39
\\
Synthetic-RCT(\small GPT-5.2\large)  $\rightarrow$ ModernBERT$_r$ & \textbf{0.2601} & \textbf{0.1722} & \textbf{0.1359} & \textbf{0.3820} & \textbf{0.3026} & \textbf{73.08} & \textbf{58.3} & \textbf{68.02}
\\
\midrule
\textit{Gold RCT} $\rightarrow$ \textit{ModernBERT$_r$} & \textit{0.2449}& \textit{0.1574} & \textit{0.2342} & \textit{0.4844} & \textit{0.3890} & \textit{73.40} & \textit{61.97} & \textit{68.70} 
\\
 \bottomrule
\end{tabular}
}
\caption{In-domain effect-size forecasting results on Test$_{id}$ for specific (level-0) queries. We report our baselines in \textit{italic}, and use \textbf{bold} to highlight the best scores among our comparison systems.}
\label{tab:query_lv1_results}
\end{table*}



\subsection{Models Evaluated}
\label{sec:exp-models}

\paragraph{Baselines.}
We first include a simple \textit{mean-effect} baseline that predicts the average effect size observed in the training data. We also test the performance of a simple \textit{retrieval-lookup} into the training set using a retrieval method to check whether predicting effect size by looking at the most similar RCT in the training corpus is beneficial. Specifically, for these experiments, we use a simple BM25 retriever that, for each query in input, answers with a simple lookup on the training and validation set. Finally, we include an upper bound for the Synthetic-RCT pipeline, i.e. \textit{Gold RCT}, which represents the performance of supervised systems starting from gold RCT descriptions.

\paragraph{LLM prompting.}
We first benchmark large language models by prompting them with queries, the simplest out-of-the-box solution available. 
Specifically, we evaluate the performance of Gemini~2.5~Flash~\cite{comanici2025gemini25pushingfrontier}, GPT-OSS~120B~\cite{agarwal2025gpt}, and GPT-5.2.\footnote{\scriptsize\url{https://openai.com/index/introducing-gpt-5-2}}
We prompt them by describing the task and adding 3 examples from the training set, using the precise description included in Appendix Table~\ref{tab:ambiguity_prompt_template}.

\paragraph{Synthetic-RCT Generators.}
To construct intermediate structured representations of queries, we employ 
GPT-5.2, alongside GPT-OSS 20B~\cite{agarwal2025gpt}, an open-weight model that serves as a low-cost alternative. 

\paragraph{Supervised Regressor.}
For supervised models, we opted for an architecture that balances accuracy and simplicity, providing the robustness required for our task while remaining efficient and easy to finetune. Specifically, we train a pretrained transformer encoder, namely ModernBERT-large~\cite{warner-etal-2025-smarter}, with a regression head on top. The model predicts the treatment effect $\hat{E}$ along with the corresponding confidence interval bounds $(\hat{CL}, \hat{CU})$ and is trained to minimize a Mean Squared Error objective over predicted values.
We train this architecture in two settings: i) ModernBERT$_q$, a model trained to start from policy queries, and ii) ModernBERT$_r$, a model trained to start from the synthetic RCTs descriptions. We report additional training details in Appendix \ref{app:training-details}.

\section{Results}
\label{sec:results}

\begin{table*}[ht]
\centering
\resizebox{\linewidth}{!}{
\begin{tabular}{l c c c c c c c c}
\toprule
\textbf{Model} & \textbf{RMSE$\downarrow$} & \textbf{MAE$\downarrow$} & \textbf{$R^2$$\uparrow$} & \textbf{Pearson$\uparrow$} & \textbf{Spearman$\uparrow$} & \textbf{Direction $\uparrow$} & \textbf{Econ. Sign. $\uparrow$} & \textbf{Stat. Sign. $\uparrow$}
\\
\toprule

\textit{mean-effect} & \textit{0.2444} & \textit{0.181} & \textit{-0.1205} & - & - & \textit{72.32} & \textit{47.5} & \textit{43.5} \\
\textit{retrieval-lookup} & \textit{0.4084} & \textit{0.2588} & \textit{-2.1269} & \textit{0.0279} & \textit{0.0379} & \textit{68.23} & \textit{45.23} & \textit{44.02}
\\

\hline
\rowcolor{gray!20}
\multicolumn{9}{c}{\textit{Prompted}} \\
Gemini 2.5 Flash & 0.3793 & 0.3027 & -1.6972 & 0.2026 & 0.1988 & 67.66 & 45.64 & 43.77
\\
GPT-OSS 120B & 0.4505 & 0.3936 & -2.8057 & 0.1659 & 0.1918 & 66.78 & 45.04 & 34.00
\\
GPT-5.2 & 0.2782 & 0.2089 & -0.4513 & 0.2503 & \textbf{0.252} & 66.31 & 48.99 & 57.80 
\\

\hline
\rowcolor{gray!20}
\multicolumn{9}{c}{\textit{Our models (Supervised)}} \\

ModernBERT$_q$ & 0.2578 & 0.1773 & -0.0603 & 0.1095 & 0.1138 & 71.23 & 55.70 & 58.23
\\

Synthetic-RCT(\small GPT-OSS-20B\large)  $\rightarrow$ ModernBERT$_r$& 0.231 & 0.1402 & 0.0019 & 0.1818 & 0.1589 & \textbf{74.93} & 57.12 & 63.23 
\\

Synthetic-RCT(\small GPT-5.2\large) $\rightarrow$ ModernBERT$_r$& \textbf{0.2241} &\textbf{ 0.1375} & \textbf{0.0591} & \textbf{0.265} & 0.223 & 73.33 & \textbf{57.88} & \textbf{63 .56}
\\


\midrule
\textit{Gold RCT} $\rightarrow$ \textit{ModernBERT$_r$} &   \textit{0.2211} & \textit{0.1302} & \textit{0.0628} & \textit{0.271} & \textit{0.2618} & \textit{73.95} & \textit{60.30} & \textit{63.61} \\

 \bottomrule
\end{tabular}
}
\caption{Out-of-domain effect-size forecasting on Test$_{ood}$, the non-health cross-sector split of Query2Effect, for specific (level-0) queries. Baselines are in \textit{italic}; \textbf{bold} marks the best score among comparison systems.}
\label{tab:ood}
\end{table*}

\begin{table*}[ht]
\centering
\resizebox{\linewidth}{!}{
\begin{tabular}{l c c c c c c c c}
\toprule
\textbf{Model} & \textbf{RMSE$\downarrow$} & \textbf{MAE$\downarrow$} & \textbf{$R^2$$\uparrow$} & \textbf{Pearson$\uparrow$} & \textbf{Spearman$\uparrow$} & \textbf{Direction $\uparrow$} & \textbf{Econ. Sign. $\uparrow$} & \textbf{Stat. Sign. $\uparrow$}
\\

\hline
\rowcolor{gray!20}
\multicolumn{9}{c}{\textit{Prompted}} \\
Gemini 2.5 Flash & 0.4001 & 0.3333 & -6.4528 & -0.1066 & -0.0341 & 74.32 & 48.65 & 28.38
\\
GPT-OSS 120B & 0.426 & 0.3436 & -7.4519	 & -0.1618 & -0.1335 & 74.32 & 50.0 & 33.78
\\
GPT-5.2 & 0.249 & 0.1688 & -1.887 & -0.1369 & -0.0765 & 75.68 & 52.7  & 50.0 
\\

\hline
\rowcolor{gray!20}
\multicolumn{9}{c}{\textit{Our models (Supervised)}} \\

ModernBERT$_q$ & 0.1872 & 0.1280 & -1.7377 & -0.0296 & -0.0806 & 71.08 & 51.70 & 66.23 
\\

Synthetic-RCT(\small GPT-OSS-20B\large)  $\rightarrow$ ModernBERT$_r$& 0.1520 &\textbf{ 0.1026} & \textbf{-0.1222} & 0.2233 & 0.1839 & 84.72 & 52.78 & 69.44
\\

Synthetic-RCT(\small GPT-5.2\large) $\rightarrow$ ModernBERT$_r$& \textbf{0.1480} & 0.1055  & -0.1332 & \textbf{0.2532} & \textbf{0.2039} & \textbf{85.72} & \textbf{54.80} & \textbf{70.22}
\\


 \bottomrule
\end{tabular}
}
\caption{Out-of-domain effect-size forecasting on AidGrade$_{ood}$, the aggregated meta analysis benchmark adapted from \citet{vivalt2020much}. Baselines are in \textit{italic}; \textbf{bold} marks the best score among comparison systems.}
\label{tab:vi}
\end{table*}

\subsection{In-domain Effect size Forecasting}

Table~\ref{tab:query_lv1_results} reports in-domain performance on effect-size forecasting from specific queries. We first note that even forecasting from \textit{gold} descriptions reaches only $R^2 \approx 0.23$, indicating that effect sizes are intrinsically difficult to predict from text and establishing a ceiling for the task. Prompting results vary widely across LLMs, with GPT-5.2 the strongest, outperforming Gemini 2.5 Flash and GPT-OSS 120B on most metrics. Notably, both the retrieval baseline and the prompted LLMs are outperformed on error-based metrics by the simple \textit{mean-effect} baseline, reflecting the strong central tendency of effect sizes in large RCT repositories \cite{vivalt2020much}.

Supervised models trained on our dataset consistently outperform direct LLM predictions. Our best supervised version, starting from synthetic RCTs, achieves a MAE of 0.1722 ($-27\%$ vs.\ GPT-5.2) and recovers most of the achievable performance, reaching $R^2 = 0.136$ and Pearson $= 0.38$ against the \textit{Gold RCT} ceiling of $0.234$ and $0.48$. Both supervised variants improve across most metrics, including RMSE, Pearson, and $R^2$, indicating stronger regression capabilities. This advantage also extends to policy-oriented metrics, with finetuned models more accurately predicting the direction of the effect and identifying economically and statistically significant results.

We observe only small differences between the two supervised strategies in-domain. Since both are trained on Query2Effect, and in-domain queries already leave little headroom below the Gold RCT ceiling, the intermediate synthetic representation yields only marginal gains here; as we show in the next section, its benefit emerges out-of-domain, where it normalizes unfamiliar phrasing toward the training distribution. When high-quality in-domain data is available, a direct regression model thus already achieves strong performance while remaining simple and efficient.

We reserve Appendix~\ref{app-add-results} for additional results highlighting the Synthetic-RCT pipeline's advantage on unspecific and abstract queries, along with qualitative examples.


\subsection{Out-of-domain Evaluation}

\paragraph{Test$_{ood}$.} Table~\ref{tab:ood} reports results on Test$_{ood}$, the out-of-domain portion of Query2Effect. In contrast to the in-domain setting, the Synthetic-RCT pipeline shows a clear advantage, improving MAE by up to $0.0398$ ($-22.4\%$) over ModernBERT$_q$. The gaps over out-of-the-box LLMs are larger still: $-0.0714$ MAE ($-34.2\%$) versus GPT-5.2, and up to $-0.26$ MAE ($-65.1\%$) versus the smaller LLMs. The pipeline also sits closer to the \textit{Gold RCT} ceiling than it does in-domain, and improves more over the \textit{mean-effect} baseline than in-domain. 

Together, these results indicate that the synthetic representation is particularly beneficial for cross-domain transfer, going beyond purely data-driven correlation toward better semantic generalization.

\paragraph{AidGrade$_{ood}$.} Table~\ref{tab:vi} reports results on the aggregate out-of-domain dataset AidGrade$_{ood}$, where the Synthetic-RCT pipeline again proves superior. 
It lowers MAE by up to $0.023$ ($-17.6\%$) over ModernBERT$_q$ and improves statistical-significance classification by $+4\%$, reflecting better confidence-interval estimation. 
The gains over out-of-the-box LLMs are even larger: $-0.063$ MAE ($-37.5\%$) versus GPT-5.2, and up to $-0.24$ MAE ($-69.3\%$) versus the smaller LLMs. 

These results, computed on the aggregated effect size benchmark, reinforce that synthetic RCT representations improve generalization on cross-sector questions and have better capabilities of dealing with abstract questions.

\section{Conclusion}\label{sec:conclusions}
We introduce the task of continuous causal effect forecasting from natural language queries and introduce Query2Effect, a new dataset derived from a large corpus of randomized controlled trials (RCTs). 
We benchmark a range of modeling approaches, including statistical and retriever-based baselines, end-to-end generative methods, and fine-tuned encoder models. 
Our results show that model scale plays an important role for generative systems, while finetuning substantially improves predictive performance. 

We further demonstrate that introducing a structured semantic abstraction via the Synthetic-RCT pipeline improves robustness and is particularly effective under domain shift.
These findings suggest that separating semantic interpretation from numerical estimation is a promising direction for modeling causal effects from text. 

We hope our work can provide a useful foundation for future research at the practically relevant intersection of natural language understanding and empirical science.

\section{Limitations}
Our experimental setup abstracts away from some of the aspects of real-world decision making and policy reasoning by relying on textual queries. 
In practice, accurate treatment effect estimation often requires rich contextual information about populations, implementation details, and outcome measurement, which can be absent in abstract and unspecific queries.
For this reason, in our experiment section, we focus on specific queries, which often contain precise and detailed information, and present additional setups in the appendix to evaluate abstract queries on averaged effects. We believe this represents a first step towards a more comprehensive analysis of this challenging and practically relevant task, and future work should focus on extending our controlled setting into a more comprehensive decision-support benchmark.

A further limitation stems from the nature of the target itself. Treatment effects are highly variable and context-dependent, yet standardized RCT effect distributions also exhibit strong central tendency, with many interventions clustering around similar magnitudes. Together, these properties limit learnability and make precise effect size prediction difficult even for in-domain examples. This is a well-known characteristic of large RCT repositories \cite{vivalt2020much, dellavigna2022rcts}, and is directly derived from the underlying distribution of our original corpus.
As a result, standard regression metrics such as $R^2$ remain relatively low, while simple mean-effect baselines can achieve competitive performance due to the strong central tendency of standardized RCT effect distributions. Consequently, improvements from query semantics are often better reflected in policy-oriented metrics, such as Pearson/Spearman correlation and directional or significance accuracy, than in absolute-error metrics alone.

Finally, our experiments focus exclusively on English queries. While the proposed framework is language agnostic, future work should evaluate its performance across other languages and contexts.

\section*{Acknowledgements}
The authors are grateful for financial support from the Knowledge for Change (KCP) Program,  administered by the World Bank, and currently funded by The Swedish International Development Cooperation Agency (SIDA), Agence Française de Développement (AFD) - French Development Agency, the Government of Japan, and the European Union.

\bibliography{anthology,custom}

\appendix

\section{Underlying RCT Database}
\label{app-rct}
Our dataset is derived from a World Bank RCT database.\footnote{\url{https://impactai.worldbank.org/blog/extracting-experimental-data-from-studies}} 
This resource aggregated evidence from randomized controlled trials, extracted through an expert-validated high-quality method, evaluating development and public policy interventions across domains such as education, health, agriculture, and social protection.

In the RCTs selected for our studies, participants are randomly assigned to a \textit{treatment group}, which receives the intervention, or to a \textit{control group}, which does not. 
Randomization ensures that differences in observed outcomes between the two groups can be causally attributed to the intervention. 
Therefore, each experiment provides an estimate of the causal impact of a specific \textit{intervention} on a measurable \textit{outcome}. 
Examples of interventions include programs such as deworming campaigns, conditional cash transfers, or new diagnostic technologies, while outcomes may include variables such as school attendance, health indicators, or household income.

We use original information about intervention and outcome descriptions.
In the original data, this also includes multiple pieces of information that are crucial to estimate effect size, such as population, context, and study design, setting, time horizon, implementation, measurement, and often sample size.

To enable comparison across heterogeneous studies, the database reports \textit{standardized effect sizes}. 
These quantities measure the magnitude of the intervention's impact relative to the variability of the outcome, allowing results from different experiments to be placed on a common scale. 

In this dataset, standardized effect sizes are all reported using Hedges' $g$, a bias-corrected standardized mean difference commonly used in meta analysis. 
Hedges' $g$ measures the difference between the mean outcome in the treatment group ($\bar{X}_T$) and the control group ($\bar{X}_C$), normalized by the pooled standard deviation $s_p$:

\[
g = J \cdot \frac{\bar{X}_T - \bar{X}_C}{s_p}
\]

where $J$ is a small-sample correction factor that reduces bias when the number of observations is limited. 
Compared to related measures such as Cohen's $d$, Hedges' $g$ provides a more reliable estimate when sample sizes are small, which is common in many field experiments.

In addition to the effect size estimate $ES_i$, the database provides the corresponding confidence interval $[CL_i, CU_i]$, which captures the statistical uncertainty surrounding the estimate. 
Together, these values allow researchers to assess both the magnitude and the statistical significance of policy effects.

\begin{table*}[ht!]
\centering
\small
\begin{tabular}{ccccl}
\toprule
\textbf{Specificity} & \textbf{E} & \textbf{A} & \textbf{U} & \textbf{Description} \\
\midrule
Query Level-0 & I0 & A0 & U0 & Fully specified query explicitly mentioning the intervention and outcome. \\
Query Level-1 & I1 & A1 & U1 & Mild paraphrasing or abstraction while preserving causal clarity. \\
Query Level-2 & I2 & A2 & U2 & Conceptually abstract phrasing describing broader intervention categories. \\
Query Level-3 & I3 & A3 & U3 & Very general policy question with minimal explicit causal structure. \\
\bottomrule
\end{tabular}

\caption{Canonical query specificity level profiles used for automatic query generation.}
\label{tab:query_profiles}

\end{table*}

\section{Prompts and configurations}
\label{app:prompts}

\subsection{Query Generation}
\label{app:prompts-query}

To operationalize the taxonomy introduced in Section~\ref{sec:dataset-taxonomy}, we define four canonical specificity profiles that span different configurations of implicitness (E), abstraction (A), and ambiguity (U). 
We use Table \ref{tab:query_profiles} to represent the distinct types of policy queries, and Table \ref{tab:ambiguity_prompt_template} for full prompt details. These profiles are used during query generation to guide the linguistic variation while preserving the causal semantics of the original RCT estimate.

Each intervention–outcome pair is associated with one query for each specificity profile, resulting in four semantically related questions that vary only in linguistic specificity. 
This design allows us to evaluate model robustness to different formulations of the same causal question while holding the underlying treatment

\subsection{Synthetic-RCT Generation}
\label{app:prompts-synth}

We use Table \ref{tab:query_to_synthrct_prompt_template} for full synthetic-RCT prompt details. 
Given a query as an input, the instruction asks to generate SyntheticRCT structure consisting of a pair of intervention description and outcome description. The rule on description is that it must be detailed (i.e. not keyword) and allows to contain or infer certain types of information (e.g. target population, when explicitly mentioned). This intermediate, possibly richer, structure allows the prediction of effect size.

\subsection{Effect-size Forecasting}
\label{app:prompts-es}

We use Table \ref{tab:inference_prompt_template} for effect-size forecasting prompt details. Given a query as input, the instruction asks to forecast the effect size and confidence intervals. Three examples are provided to guide the inference, one for each possible statistical significance class.

\section{AidGrade Dataset Details}
\label{app:ood}
As an out-of-domain benchmark, we include a secondary benchmark based on the meta analysis dataset AidGrade \citet{vivalt2020much}. 
This dataset aggregates evidence from more than 600 RCTs across development economics interventions in multiple sectors, such as education and farming. 
The original resource reports aggregate estimates for a set of intervention-outcome pairs, including information as (i) Intervention name (e.g., \textit{Deworming},  \textit{Conditional Cash Transfers}), (ii) Outcome name (e.g., \textit{School attendance}, \textit{Height}), and (iii) standardized effect size.
.

\noindent 
After filtering entries with incomplete information, the resulting benchmark contains 73 intervention-outcome pairs that, although smaller than our out-of-domain dataset, for each estimate summarizes evidence from multiple RCTs reflecting a stable aggregate effect. 
We then convert each intervention-outcome pair into an experimental question of the form:
\textit{"What is the impact of [intervention] on [outcome]?"}.

\noindent
This produces a compact set of natural-language queries that closely resembles our abstract queries, i.e., those created with specificity level-3. 
This enables us to evaluate whether models can generalize causal effect predictions to previously unseen, abstract formulations with aggregated evidence.



\section{Evaluation}
\label{app-metrics}
This section provides formal definitions for all evaluation metrics used in the main paper.

\subsection{Regression Metrics}

We measure numerical accuracy by comparing predicted effect sizes $\hat{ES}_i$ with ground-truth estimates $ES_i$ using standard regression metrics.

\paragraph{Root Mean Squared Error (RMSE).}
RMSE measures the average squared deviation between predicted and true effect sizes:

\[
RMSE = \sqrt{\frac{1}{N}\sum_{i=1}^{N}(ES_i - \hat{ES}_i)^2}.
\]

It penalizes large errors more strongly and therefore reflects sensitivity to large prediction deviations.

\paragraph{Mean Absolute Error (MAE).}
MAE measures the average absolute difference between predicted and true effects:

\[
MAE = \frac{1}{N}\sum_{i=1}^{N}|ES_i - \hat{ES}_i|.
\]

Compared to RMSE, MAE provides a more robust estimate of typical prediction error.

\paragraph{$R^2$ (Coefficient of Determination).}
$R^2$ measures the proportion of variance in the true effect sizes explained by the model:

\[
R^2 = 1 - \frac{\sum_{i}(ES_i - \hat{ES}_i)^2}{\sum_{i}(ES_i - \bar{ES})^2}.
\]

Higher values indicate that predictions better capture variation in treatment effects.

\paragraph{Pearson Correlation.}
Pearson correlation evaluates the linear relationship between predicted and true effect sizes:

\[
\rho_P = \frac{\sum_i (ES_i - \bar{ES})(\hat{ES}_i - \overline{\hat{ES}})}
{\sqrt{\sum_i (ES_i - \bar{ES})^2}\sqrt{\sum_i (\hat{ES}_i - \overline{\hat{ES}})^2}}.
\]

\paragraph{Spearman Correlation.}
Spearman correlation measures rank correlation between predictions and ground truth. 
It evaluates whether models preserve the relative ordering of treatment effects, regardless of exact numerical magnitudes

\subsection{Policy-Oriented Metrics}

\paragraph{Direction Accuracy}

Direction accuracy measures the proportion of cases where predicted and true effects share the same sign:

\[
\text{Direction} =
\frac{1}{N}\sum_{i=1}^{N}
\mathbf{1}[\text{sign}(\hat{ES}_i) = \text{sign}(ES_i)]
\]

\paragraph{Economic Significance}

Following common conventions in development economics and policy evaluation, we consider an effect economically meaningful when $|ES_i| > 0.1$. This threshold reflects effect sizes that are typically considered practically relevant for large-scale policy interventions, as documented in meta analyses of social and development programs.\footnote{\scriptsize\url{https://blogs.worldbank.org/en/impactevaluations/notes-aeas-present-bias-20-years-should-we-give-sds-effect-size}}

Predictions are classified accordingly and evaluated using classification accuracy.

\paragraph{Statistical Significance Classification}

We classify each estimate using the 95\% confidence interval $(CL_i, CU_i)$:

\begin{itemize}
\item Positive: $CL_i > 0$
\item Negative: $CU_i < 0$
\item Non-significant: $CL_i \leq 0 \leq CU_i$
\end{itemize}

Model predictions are evaluated using the micro-averaged F1 score across these three classes.
Since statistical significance depends on study-specific factors such as sample size and variance, our predictions implicitly assume the distribution of experimental conditions observed in the underlying RCT corpus. 
This measure is particularly important for evaluating confidence bounds, a key output of our predictions, which reflect whether an intervention would likely be statistically distinguishable from zero under typical RCT settings.

\section{Training Details}
\label{app:training-details}

Our trained models predict the treatment effect $\hat{E}$ along with the corresponding confidence interval bounds $(\hat{CL}, \hat{CU})$. 
They are trained to minimize a Mean Squared Error objective over all predicted quantities, such as:

$$
\mathbb{E}=\left[(E-\hat{E})^2+(CL-\hat{CL})^2 + (CU - \hat{CU})^2 \right]
$$

Our systems are implemented using the PyTorch Lightning framework\footnote{\url{https://lightning.ai/}} and the HuggingFace Transformers library\footnote{\url{https://huggingface.co/docs/transformers}}. 
Training is performed on a single RTX~4090 GPU using the AdamW optimizer \cite{loshchilov2017decoupled} with a learning rate of $5\times10^{-6}$ and a linear scheduler with a warm-up of 10\% of total training steps. Models are trained for up to four epochs with early stopping (patience of 20) and validation every 30\% of the steps per epoch, resulting in approximately 6 hours of training per model.

\section{Additional Results and Analysis}
\label{app-add-results}

\subsection{Providing Additional Training Information}
We test our prompted systems in an additional setup, in which we explicitly include in the prompt some additional training statistics, such as standard deviation and training mean effect size. We do so to benchmark whether this information can help close the gap with supervised systems that learn these parameters during finetuning. 
In Table \ref{tab:query_lv1_resultss} we show our results, highlighting a marginal improvement on large models. 
Smaller models instead show a greater improvement, reducing prediction errors and slightly improving rank correlations.

\begin{table*}[ht!]
\centering
\resizebox{\linewidth}{!}{
\begin{tabular}{l c c c c c c c c}
\toprule
\textbf{Model} & \textbf{RMSE$\downarrow$} & \textbf{MAE$\downarrow$} & \textbf{$R^2$$\uparrow$} & \textbf{Pearson$\uparrow$} & \textbf{Spearman$\uparrow$} & \textbf{Dir. $\uparrow$} & \textbf{Econ. Sign. $\uparrow$} & \textbf{Stat. Sign. $\uparrow$}
\\

\toprule
\textit{mean-effect} & \textit{0.2788} & \textit{0.1946} & \textit{-0.0797} & - & - & \textit{73.02} & \textit{48.46} & \textit{45.3} \\
\textit{retrieval-lookup} & \textit{0.3880} & \textit{0.2526} & \textit{-0.9235} & \textit{0.2423} & \textit{0.164} & \textit{67.8} & \textit{51.0} & \textit{53.3}
\\

\hline
\rowcolor{gray!20}
\multicolumn{9}{c}{\textit{Zero-shot + 3 examplesed}} \\
Gemini 2.5 Flash & 0.4503 & 0.3542 & -1.5908 & 0.1711 & 0.2014 & 59.36 & 51.36 & 41.37
\\
GPT-OSS 120B & 0.4900 & 0.4160 & -2.0682 & 0.1977 & 0.2213 & 59.80 & 48.88 & 32.98
\\
GPT-5.2 & 0.3165 & 0.2374 & -0.2801 & 0.2700 & 0.3008 & 65.23 & 52.33 & 62.22 
\\

\hline
\rowcolor{rowgray}
\multicolumn{9}{c}{\textit{Zero-shot + 3 examples + Training Info}} \\
Gemini 2.5 Flash & 0.3982 & 0.3201 & -1.0254 & 0.151 & 0.1956 & 60.86 & 50.24 & 36.28
\\
GPT-OSS 120B & 0.3693 & 0.3002 & -0.7421 & 0.1643 & 0.1925 & 63.48 & 48.84 & 36.03 
\\
GPT-5.2 & 0.3062 & 0.2195 & -0.1977 & 0.2373 & 0.273 & 62.03 & 51.79 & 64.69 
\\

\hline
\rowcolor{gray!20}\multicolumn{9}{c}{\textit{Our models (Supervised)}} \\

ModernBERT$_q$ & 0.2659 & 0.1743 & 0.097 & 0.348 & 0.2988 & 71.82 & 57.71 & 66.78 
\\
Synthetic-RCT(\small GPT-OSS-20B\large) $\rightarrow$ ModernBERT$_r$ & 0.2685 & 0.179 & 0.0807 & 0.3349 & 0.3002 & 72.63 & 57.71 & 65.39
\\
Synthetic-RCT(\small GPT-5.2\large)  $\rightarrow$ ModernBERT$_r$ & \textbf{0.2601} & \textbf{0.1722} & \textbf{0.1359} & \textbf{0.3820} & \textbf{0.3026} & \textbf{73.08} & \textbf{58.3} & \textbf{68.02}
\\
Synthetic-RCT(\small Gemini-3-Pro\large) $\rightarrow$ ModernBERT$_r$ & 0.2704 & 0.1812 & 0.0671 & 0.3134 & 0.2714 & 73.01 & 55.53 & 65.39
\\
\midrule
\textit{Gold RCT} $\rightarrow$ \textit{ModernBERT$_r$} & \textit{0.2449}& \textit{0.1574} & \textit{0.2342} & \textit{0.4844} & \textit{0.3890} & \textit{73.4} & \textit{61.97} & \textit{68.7} 
\\
 \bottomrule
\end{tabular}
}
\caption{In-domain Effect-size forecasting results on Test$_{id}$ for specific queries, i.e., those with specificity level-0. We report results of our baselines in \textit{italic}, and use \textbf{bold} to highlight the best performance among our comparison systems.}
\label{tab:query_lv1_resultss}
\end{table*}

\subsection{Results on Unspecific Queries}
In our main results, we focus on describing the performance of our comparison systems on specific queries, as they are created to be faithful to descriptions of RCT, containing contextual information such as population, country, and implementation, inter alia, which are crucial to estimate a precise effect size.
In the following, we focus on queries with lower specificity (e.g., level-1, level-2, and level-3) that mimic more practical and real-world scenarios.

\begin{figure}[t!]
    \centering
    \includegraphics[width=0.98\linewidth]{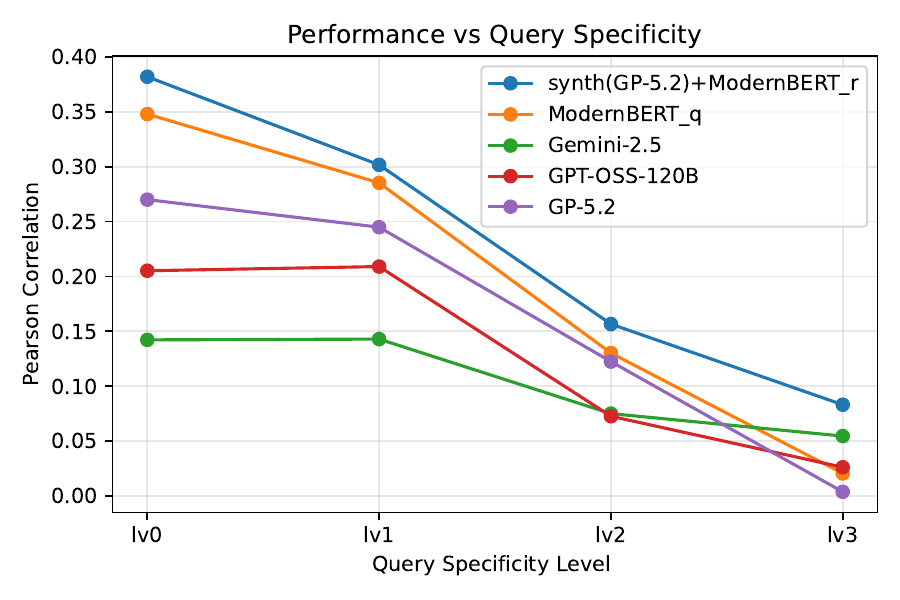}
    \caption{Performance degradation at decreasing levels of specificity in terms of Pearson correlation on Test$_{id}$.}
    \label{fig:pearson_vs_difficulty}
\end{figure}

\paragraph{Robustness to Different Levels of Specificity}
To evaluate robustness to linguistic variation, in Figure \ref{fig:pearson_vs_difficulty}, we analyze performance across different levels of query specificity and absence of information. 
While these queries have a less strict prediction target, as effect size prediction depends on the above-mentioned factors, this setting allows us to assess differences between models when these details are missing, which can happen in practical scenarios.
We observe that, as query specificity decreases and information is less representative of full treatment details, model performance also degrades. 
However, our Synthetic-RCT pipeline performs the best across all query specificity levels despite suffering from the same degradation.

\paragraph{Evaluating Unspecific Queries with Average Effects}
To further evaluate the performance of our systems on unspecific queries, we perform an additional test by averaging the effects of unspecific queries. 

Queries level-3 are designed to only contain superficial information about intervention and outcome descriptions, and therefore their target is loosely correlated with the underlying RCT-specific effect-size. 
Instead, similarly to the setting shown in AidGrade$_{ood}$, they are similar to general meta analysis questions which often report average effects of multiple RCTs that fall under the description of the question.

To properly test performance in this setting, we modify the target of unspecific queries to match the average value of similar estimates.
To find similar estimates, we use two additional estimate-level variables that are present in the World Bank corpus, \textit{"intervention name"} and \textit{"outcome name"}, names that identify both interventions and outcomes. 
Specifically, we test model performances of the 2,062 level-3 queries in Test$_{id}$ on the average effect of those that matched both intervention and outcome name. 
We report that 1,804 of those queries have at least one similar estimate, and, on average, every query has 2.7 similar estimates.
We report the results of this additional test in Table \ref{tab:oodddddd}, showing that the performance of our compared systems improves, and becomes closer to specificity level-0 query results. 
Interestingly, we show a substantial improvement compared to measuring error without averaged effects, showing that this setting provides a better interpretation of abstract query results, and should be prioritized by future work testing performance on abstract queries.

\begin{table*}[ht]
\centering
\resizebox{\linewidth}{!}{
\begin{tabular}{l c c c c c c c c}
\toprule
\textbf{Model} & \textbf{RMSE$\downarrow$} & \textbf{MAE$\downarrow$} & \textbf{$R^2$$\uparrow$} & \textbf{Pearson$\uparrow$} & \textbf{Spearman$\uparrow$} & \textbf{Direction $\uparrow$} & \textbf{Econ. Sign. $\uparrow$} & \textbf{Stat. Sign. $\uparrow$}
\\
\toprule

\textit{mean-effect} & \textit{0.2444} & \textit{0.181} & \textit{-0.1205} & - & - & \textit{72.32} & \textit{47.5} & \textit{43.5} \\
\textit{retrieval-lookup} & \textit{0.5184} & \textit{0.3488} & \textit{-2.3358} & \textit{0.0252} & \textit{-0.0122} & \textit{67.23} & \textit{45.33} & \textit{43.02}
\\

\hline
\rowcolor{gray!20}
\multicolumn{9}{c}{\textit{Prompted}} \\

Gemini 2.5 Flash & 0.3842 & 0.3227 & -1.753 & 0.1768 & 0.1566 & 69.21 & 48.39 & 35.77
\\
GPT-OSS 120B & 0.4602 & 0.4081 & -2.9488 & 0.1894 & 0.1753 & 69.33 & 47.38 & 24.00
\\

\hline
\rowcolor{gray!20}
\multicolumn{9}{c}{\textit{Our models (Supervised)}} \\

ModernBERT$_q$ & 0.2756 & 0.1735 & -0.1003 & 0.1781 & 0.1737 & 70.03 & 47.38 & 66.81
\\

Synthetic-RCT(\small GPT-OSS-20B\large)  $\rightarrow$ ModernBERT$_r$& 0.2677 & 0.1636 & 0.122 & 0.2200 & 0.2133 & \textbf{71.22} & 49.88 & 66.22 
\\



\hline

\rowcolor{gray!20}
\multicolumn{9}{c}{\textit{Our models (Supervised) - non averaged}} \\
\textit{ModernBERT$_q$ } & \textit{0.3173} & \textit{0.2317} & \textit{-0.2863} & \textit{0.0831} & \textit{0.0833} & \textit{73.08} & \textit{49.71} & \textit{57.66}
\\

\textit{Synthetic-RCT(\small GPT-OSS-20B\large)}  $\rightarrow$ ModernBERT$_r$& \textit{0.2890} & \textit{0.2048} & \textit{-0.0649} & \textit{0.1309} & \textit{0.1299} & \textit{70.77} & \textit{53.55} & \textit{60.78}
\\

 \bottomrule
\end{tabular}
}
\caption{Evaluation of our comparison systems on the averaged effects of similar queries in Test$_{id}$ when dealing with specificity level-3 queries. On bottom, we include the results of some of our comparison systems with the original non-averaged effect size.}
\label{tab:oodddddd}
\end{table*}

\subsection{Confidence Interval Performance}
We evaluate the performance of our models to predict confidence intervals using statistical significance classification accuracy. 
While this does not directly evaluate the accuracy of confidence intervals, it evaluates their most important implications on study impact: the capabilities of determining whether this is statistically significant or not. 
From Table \ref{tab:query_lv1_resultss}, we see that our supervised models have better performance on statistical significance classification, which is directly impacted by their superiority in predicting confidence intervals.

Additionally, we report information about the validity of confidence interval predictions: all our comparison systems have produced valid outputs in which, for estimate $EST_i$, confidence bounds respect the constraint of $CL_i < ES_i < CU_i$.

We also highlight that those intervals depend on study-specific factors such as sample size and variance, and therefore, our predictions implicitly assume the distribution of experimental conditions observed in the underlying RCT corpus.
For this reason, the average width of the predicted confidence interval is not a meaningful parameter to evaluate uncertainty. 
We only report the high variance and poor correlation of prompted systems, which do not have information about the underlying corpus.

\subsection{Qualitative Error Analysis}
\label{app:qualitative}
To better understand the model limitations, we analyze prediction errors across intervention categories. 
Table~\ref{tab:syntheticrct_examples} shows representative predictions from the Synthetic-RCT pipeline. 
We report three best predictions (near-zero error) and three worst cases with large deviations between the predicted and the gold effect sizes.

We can see that large positive and negative effect sizes are more difficult to predict than moderate effects, suggesting that rare outcomes remain challenging for both supervised models and LLM baselines.

\subsection{Can we Automatically Predict effect sizes?}
Our experiments on continuous causal forecasting from textual queries demonstrate that automatic systems can produce meaningful effect-size estimates.
Our best model achieves MAE = 0.17 in-domain, while prior work on human forecasting reports MAE = 0.24 on a related treatment-effect prediction setting \cite{dellavigna2025forecasting}. 
We also observe the same mild overestimation bias in automatic systems: whereas humans in prior work overestimate treatment effects by approximately +0.11 on average \cite{dellavigna2025forecasting}, our Synthetic-RCT pipeline shows a smaller average bias of +0.08. 
Although these evaluations are not directly comparable due to differences in datasets and query formats, the comparison provides a useful qualitative reference point for the difficulty of the task. 
Future work should directly compare expert and AI predictions in treatment-effect forecasting.

\begin{table}[t]
\centering

\scriptsize
\begin{tabular}{p{0.47\textwidth}}
\hline
\textbf{Prompt} \\[0.5em]

\begin{minipage}{0.45\textwidth}
\begin{verbatim}
You are an expert dataset generator for causal inference and
development policy research.

Your task is to generate FOUR distinct natural-language causal
queries about a randomized controlled trial (RCT), based on the
structured information below.

You must strictly follow these rules:
1. Generate exactly FOUR queries.
2. Each query must be ONE sentence.
3. Do NOT include answers, explanations, metadata, or formatting
   beyond what is requested.
4. Queries must differ meaningfully in implicitness, abstraction,
   and ambiguity.
5. Do NOT hallucinate details that are not present in the RCT
   (e.g., dosage, sample size, metrics).
6. When information is not allowed by the difficulty level,
   it must be omitted, not guessed.
7. Queries should sound like realistic questions asked by
   policymakers, practitioners, or researchers.

--------------------------------------------------
RCT INFORMATION
--------------------------------------------------
Intervention:
- Description: {intervention_description}

Outcome:
- Description: {outcome_description}

Sector:
- Description: {sector}

--------------------------------------------------
SPECIFICITY TAXONOMY
--------------------------------------------------
Implicitness:
- I0: All causal elements explicit
- I1: Population or comparator implicit
- I2: Outcome abstracted or generalized
- I3: Most elements implicit

Abstraction:
- A0: Concrete phrasing
- A1: Paraphrased but concrete
- A2: Conceptual abstraction
- A3: High-level framing

Ambiguity:
- U0: Clear causal intent
- U1: Mild underspecification
- U2: Multiple plausible interpretations
- U3: Ill-posed or vague

--------------------------------------------------
QUERY GENERATION INSTRUCTIONS
--------------------------------------------------
Generate the following queries, each corresponding
to one difficulty profile:
1. I0-A0-U0 (Fully explicit, concrete, unambiguous)
2. I1-A1-U1 (Implicit elements, paraphrased,
   mildly underspecified)
3. I2-A2-U2 (Conceptual abstraction with
   multiple plausible interpretations)
4. I3-A3-U3 (Very high-level,
   ill-posed causal question)

--------------------------------------------------
OUTPUT FORMAT (STRICT)
--------------------------------------------------
Each object must have the following structure:
{
  "query": "<one-sentence>",
  "difficulty": {
    "implicitness": "I0|I1|I2|I3",
    "abstraction": "A0|A1|A2|A3",
    "ambiguity": "U0|U1|U2|U3"
  }
}
Do NOT include any additional text outside the JSON.
\end{verbatim}
\end{minipage}

\\
\hline
\end{tabular}
\caption{Prompt Template for generating four queries with varying scale of implicitness-abstraction-ambiguity.}
\label{tab:ambiguity_prompt_template}
\end{table}

\begin{table}[t!]
\centering

\scriptsize
\begin{tabular}{p{0.47\textwidth}}
\hline
\textbf{Prompt} \\[0.5em]

\begin{minipage}{0.45\textwidth}
\begin{verbatim}
You are an expert in causal inference, randomized controlled 
trials (RCTs), and development policy evaluation.

Your task is to predict a natural-language causal QUERY into a 
SYNTHETIC RCT STRUCTURE, specifically a pair of 
intervention description and outcome description.

The goal is NOT to invent a full study, but to extract or 
cautiously infer the minimal structured information (here: 
intervention and outcome descriptions) that can reasonably be 
derived from the query.

------------------------------------------------------------
CORE PRINCIPLES (STRICT)
------------------------------------------------------------
1. Generate a detailed description (rather than just a keyword 
or keyphrase) for each of the intervention and outcome.
2. Prefer underspecification over extrapolating knowledge 
without basis.

------------------------------------------------------------
WHAT YOU MAY INFER
------------------------------------------------------------
You MAY extract or cautiously infer:
- Intervention type (if named or clearly implied)
- Outcome variable (possibly abstracted)
- Target population (if explicit)
- Dosage, intensity, or implementation details (if explicit)
- Broad geographic region or income level (if explicitly 
mentioned)

------------------------------------------------------------
INPUT
------------------------------------------------------------
QUERY: "{query}"

------------------------------------------------------------
OUTPUT SCHEMA (STRICT)
------------------------------------------------------------
Return a JSON object with the following structure:
{
  "intervention": <string or null>,
  "outcome": <string or null>
}

------------------------------------------------------------
IMPORTANT CONSTRAINTS
------------------------------------------------------------
- Use null rather than vague placeholders.
- Do not add fields not listed in the schema.
- Keep descriptions detailed, following the given examples.
- Ensure internal consistency across fields.

Return ONLY the JSON object.
Do NOT include explanations or commentary.
\end{verbatim}
\end{minipage}

\\
\hline
\end{tabular}
\caption{Prompt Template for Synthetic RCT generation.}
\label{tab:query_to_synthrct_prompt_template}
\end{table}

\begin{table}[t]
\centering

\scriptsize
\begin{tabular}{p{0.47\textwidth}}
\hline
\textbf{Prompt} \\[0.5em]

\begin{minipage}{0.45\textwidth}
\begin{verbatim}
You are a model that predicts the effect size in social science
experiments for forward-looking policy questions.

Given a query, your task is to produce four numbers:
1. Hedges g score: a float between -2 and 2 representing the
   estimated impact of the intervention on the outcome.
2. Lower bound of the confidence interval for Hedges g:
   must be a float smaller than Hedges g.
3. Upper bound of the confidence interval for Hedges g:
   must be a float larger than Hedges g.


Format your output exactly as a JSON object like:

{"Hedges_g": <float between -2 and 2>,
 "Hedges_g_ci_lower": <float smaller than Hedges_g>,
 "Hedges_g_ci_upper": <float larger than Hedges_g>,

 }

---
I will give you three examples of intervention–outcome pairs
with their measured Hedges g values.

Example 1 (Positive + Statistically significant):
"Does providing EMDR sessions help alleviate trauma symptoms
reported by parents for affected children?"

Output:
{"Hedges_g": 1.5956,
 "Hedges_g_ci_lower": 0.9756,
 "Hedges_g_ci_upper": 2.2156,
}

Example 2 (Negative + Statistically significant):
"How does the regular I-2 Newcastle disease vaccination affect
chick mortality from disease?"

Output:
{"Hedges_g": -1.1956,
 "Hedges_g_ci_lower": -1.6002,
 "Hedges_g_ci_upper": -0.791,
}

Example 3 (Statistically insignificant):
"How does providing immediate ART during home-based HIV testing
influence overall patient well-being and care effectiveness?"

Output:
{"Hedges_g": 0.0602,
 "Hedges_g_ci_lower": -0.1807,
 "Hedges_g_ci_upper": 0.3011,
}

-----
Now predict the Hedges g for the following example. The query is
taken from a test set of a full dataset of RCT-to-effect-size
estimations.

Training data effect size distribution:
Mean: 0.2669
Variance: 0.1847
Standard Deviation: 0.4297

Therefore most values will be close to the mean value.

Additionally, the typical (median) sample size in the training
data is 627.

QUERY:
{query}
\end{verbatim}
\end{minipage}

\\
\hline
\end{tabular}
\caption{Prompt Template for effect size prediction in the setting in which 3 examples and training information are provided. Training data effect size distributions are \textit{excluded} for out-of-domain experiments.}
\label{tab:inference_prompt_template}
\end{table}

\begin{table*}[t]
\centering
\resizebox{\linewidth}{!}{
\begin{tabular}{l p{6cm} p{13cm} l l l}
Type & Query & Synthetic RCT Description & Gold & Pred. & Error \\
\toprule

Best \#1 & Do monthly nurse home visits with patient education, self-management coaching, and physician care management lead to improved ADL scores, indicating lesser functional decline, among Medicare beneficiaries with ADL/IADL impairments compared to standard care? & \textbf{Intervention:} Monthly home visits by registered nurses to Medicare beneficiaries who have documented impairments in Activities of Daily Living (ADL) or Instrumental Activities of Daily Living (IADL). Each visit includes patient education on disease‑related self‑management, coaching to enhance patients’ ability to manage their own health care, and coordination with the beneficiaries’ physicians for ongoing care management. Visits are scheduled once per month for the duration of the study. \newline
\textbf{Outcome:} Change in ADL score over the study period, where lower scores (or larger declines) indicate less functional decline. The outcome focuses on the functional ability of Medicare beneficiaries with ADL/IADL impairments and compares post‑intervention scores to those received with standard care.
& 0.204 & 0.204 & -0.000 \\

\midrule

Best \#2 & Does offering a lottery ticket redeemable upon check-up at public health clinics reduce the prevalence of overweight status based on BMI among individuals aged 40-70 without prior cardiovascular disease in Nueva Ecija 4.5 months post-intervention, compared to a control group not receiving this incentive? & 
\textbf{Intervention:} A randomized incentive program in which eligible adults aged 40‑70 years without prior cardiovascular disease in Nueva Ecija are given a lottery ticket that can be redeemed for a prize only after completing a scheduled health check‑up at a public health clinic. At enrollment, each participant receives a unique ticket; they are instructed to present it during the clinic visit and are then entered into a draw for a predetermined reward. The intervention is delivered through community health centers, with administrators verifying ticket presentation and recording check‑up attendance over the initial 4.5‑month period. \newline
\textbf{Outcome:} The prevalence of overweight status measured by Body Mass Index (BMI$\geq$ 25 kg/m²) among participants 4.5 months after the intervention. This outcome captures the proportion of the study cohort who remain classified as overweight, allowing comparison between the incentive group and a control group that did not receive the lottery ticket.
& -0.017 & -0.017 & +0.000 \\

\midrule

Best \#3 & What is the causal effect of offering comprehensive Medicaid benefits through the Oregon Health Plan Standard to low-income, uninsured, able-bodied adults in Oregon, compared to not offering these benefits, on their self-reported average out-of-pocket medical spending 25 months after the Medicaid lottery? & 
\textbf{Intervention:} Low‑income, uninsured, able‑bodied adults in Oregon who win the Medicaid lottery are offered comprehensive Medicaid benefits through the Oregon Health Plan (OHP) Standard. The OHP Standard provides full coverage for inpatient, outpatient, prescription, mental‑health, dental, and other related medical services, with no standard cost‑sharing, for the duration of the study period. The control group receives none of these benefits. \newline
\textbf{Outcome:}The self‑reported average monthly out‑of‑pocket medical spending of participants, measured 25 months after the Medicaid lottery decision. Higher values indicate greater personal expenditure on health care.
& 0.040 & 0.039 & -0.001 \\

\toprule

Worst \#1 & Does providing nutrition education to pregnant women in their third trimester improve their mean total weight gain in kilograms during the third trimester compared to not receiving this education? & 
\textbf{Intervention:} The nutrition education program delivers structured dietary counseling and instructional materials to pregnant women during their third trimester. Women receive face‑to‑face sessions at weekly antenatal clinic visits, led by trained midwives or dietitians, covering recommended caloric intake, macro‑ and micronutrient sources, meal planning strategies, and weight‑gain monitoring. The curriculum is delivered as a combination of individual counseling and group educational talks and is offered throughout the entire third trimester. \newline
\textbf{Outcome:} Mean total weight gain in kilograms measured during the third trimester, calculated as the difference between the weight recorded at the start of the third trimester and the final weight recorded at the end of the trimester at routine antenatal visits.
& 1.959 & 0.425 & -1.534 \\

\midrule

Worst \#2 & What is the causal effect of offering female shoppers a 20\% price discount on fruits, vegetables, water, and low-calorie beverages for three months on their weekly spending on other household items like rice and tobacco at 6-month follow-up, compared to a control group? & 
\textbf{Intervention:} The intervention provides a 20\% price discount coupon to every female shopper purchasing from designated retail outlets for a period of three consecutive months. The discounted categories include fresh produce (fruits and vegetables), bottled water, and low‑calorie beverages, while the discount is not applied to other household goods such as rice or tobacco. \newline
\textbf{Outcome:} Weekly monetary spending by the treated female shoppers on other household items—specifically staples like rice and tobacco products—measured at a 6‑month follow‑up after the discount period ends.
& -1.144 & 0.123 & +1.267 \\

\midrule

Worst \#3 & Does providing home-based breastfeeding support by certified nurse lactation consultants to mothers of term and near-term newborns affect their maternal reported costs after hospital discharge up to day 7? & 
\textbf{Intervention:} Home‑based breastfeeding support delivered by certified nurse lactation consultants (CNLCs) to mothers of term and near‑term newborns during the first week after hospital discharge. Each CNLC visits the mother and infant in the home once per day for the first three days and again on days five and seven, providing individualized counseling on latch, positioning, milk supply monitoring and problem‑solving for common breastfeeding challenges. The visits include education on infant nutrition, guidance on safe formula use if needed, and referral pathways for follow‑up care. \newline
\textbf{Outcome:} Maternal self‑reported total out‑of‑pocket costs incurred between hospital discharge and day seven postpartum. These costs capture expenses for infant feeding (formula, breast‑milk supplements, feeding accessories), transportation to health facilities, childcare support, and any additional medical or social services required during the first week after birth.
& -1.254 & -0.041 & +1.213 \\

\bottomrule
\end{tabular}
}
\caption{Representative predictions from the Synthetic-RCT pipeline. The top rows show near-perfect predictions, while the bottom rows illustrate failure cases where the model misinterprets outcome scales or intervention effects.}
\label{tab:syntheticrct_examples}
\end{table*}

\end{document}